\documentclass[]{icas2026} 
\def\BibTeX{{\rm B\kern-.05em{\sc i\kern-.025em b}\kern-.08em
    T\kern-.1667em\lower.7ex\hbox{E}\kern-.125emX}}

\usepackage{xcolor}
\usepackage{graphicx}
\usepackage{tikz}
\usetikzlibrary{arrows.meta, positioning, fit, calc, backgrounds, shapes}

\title{Offline Multimodal Large Language Models for Decision Support in Air Operations}

\AuthorPaper[1]{Joao P. A. Dantas}
\AuthorPaper[1]{Jelton A. Cunha}
\AuthorPaper[1]{Gabriel Dietzsch}
\affil[1]{Instituto de Estudos Avançados, São José dos Campos, São Paulo, Brazil, 12.228-001}

\abstractEnglish{
Air operations rely on complex rules, established procedures, and time-critical analysis under limited connectivity and strict security constraints. In such environments, analysts must combine written doctrine with images, often without access to external computing resources. This paper studies offline large language models as decision support tools, deployed in isolated and restricted environments to give analysts access to doctrinal knowledge that remains traceable to its original sources, through natural language interaction. We describe a modular retrieval-augmented architecture suitable for operation without Internet connectivity, supporting both text and image input from technical manuals. As a first step toward evaluating this architecture, we report a pilot study with four image analysts of the Brazilian Air Force, combining (i) a doctrinal knowledge assessment based on their electronic-target identification doctrine, comparing human and proposed system performance on the same test, and (ii) a measurement of the cognitive workload involved in manually producing a reconnaissance target report (\textit{Relatório de Missão de Reconhecimento} --- REMIR) without AI assistance. The results show a demanding manual task, especially in terms of mental demand (6.0/7) and effort (5.0/7), while the proposed system matches the human score (8/10) and completes the assessment in 7.1 minutes (compared to a human average of 26.5 minutes), establishing a baseline for future AI-assisted evaluation. Finally, we describe a future evaluation protocol to systematically compare manual and AI-assisted workflows.
}


\begin{document}

\body 

\section{Introduction}

Modern air operations involve large amounts of information, strict rules, and time-critical decisions, placing heavy demands on operators' situation awareness and mental workload \cite{endsley2017toward,wickens2008multiple}. Command and control (C2) and intelligence, surveillance, and reconnaissance (ISR) tasks require analysts to interpret doctrine, operational guidance, and a wide range of intelligence products, often under limited connectivity and strict security rules that prevent access to external computing resources \cite{alberts2006understanding,deptula2001effects}.
 
One such intelligence product is image analysis, where analysts must compare visual features with exact details described in technical manuals that can be hundreds of pages long. When performed manually under time pressure, this comparison is mentally demanding and increases the risk of errors or poorly supported judgments \cite{parasuraman2000model,hoffman2018metrics}. Effective decision support in this setting must therefore make it faster to find reliable information, while keeping the analyst responsible and able to trace every conclusion back to its source in the doctrine \cite{sharda2018analytics}. A clear example of this challenge is electronic target intelligence: to identify a radar, antenna, or command post in a reconnaissance image, an analyst must match observed features, such as reflector size and antenna layout, with details described in doctrine, such as frequencies and functions \cite{fabmca2005}.
 
Recent progress in large language models (LLMs) has enabled natural language interaction with large collections of text, and retrieval-augmented generation (RAG) further connects model outputs to real documents, reducing the risk of fabricated content \cite{lewis2020rag,ji2023survey}. However, the most capable systems run on cloud infrastructure, which is not compatible with air-gapped intelligence environments, where transmitting classified images or doctrine over external networks is prohibited \cite{alberts2006understanding}. This gap between the capabilities of modern LLMs and the security limits of operational settings is the main motivation for this work.
 
In this work, the LLM is treated as a cognitive support tool, not as a decision maker: it retrieves relevant parts of the doctrine, explains rules and procedures, and supports the analyst's reasoning, while the final decision stays with the analyst, following well-known principles of human-centered AI \cite{amershi2019guidelines,parasuraman2000model}. Given this framing, this paper makes three main contributions:
\begin{itemize}

\item We identify the requirements for using offline LLMs as decision support tools in restricted aerospace environments, such as air-gapped operation, traceability to sources, and human accountability.

\item We propose a modular RAG architecture that meets these requirements, running fully offline and producing answers that can be traced back to their sources.

\item We report a pilot study with image analysts of the Brazilian Air Force (FAB) that compares human and AI performance on a doctrinal knowledge test and measures the cognitive workload of the manual electronic-target reporting task, establishing a baseline for future AI-assisted evaluation and guiding the design of a full within-subjects evaluation protocol.

\end{itemize}

This study is framed as a feasibility-oriented pilot, and its empirical results should be interpreted descriptively rather than as confirmatory evidence of operational effectiveness.

The rest of this paper is organized as follows. Section~\ref{sec:related} reviews related work on LLM-based decision support in defense and on multimodal retrieval-augmented generation. Section~\ref{sec:architecture} describes the proposed offline architecture, including the document processing pipeline, two-phase retrieval, and the human-in-the-loop feedback mechanism. Section~\ref{sec:casestudy} presents the case study and evaluation methodology based on FAB's doctrine manual for electronic target identification. Section~\ref{sec:results} reports the experimental results, and Section~\ref{sec:conclusion} concludes and describes future work.

\section{Related Work}
\label{sec:related}
 
Decision support in defense and aerospace has historically relied on expert systems, structured workflows, and deterministic databases \cite{alberts2006understanding,sharda2018analytics}. While such systems provide consistency, they are often rigid and have trouble handling unstructured narrative data, such as tactical debriefs, mission reports, and multi-page doctrinal manuals \cite{deptula2001effects}. Large language models introduce conversational interfaces capable of parsing natural language and producing coherent summaries \cite{lewis2020rag}. This research line has explored several related directions in this area: offline language-based support for tactical decision-making in air combat \cite{dantas2025offline}, radar-based reconstruction of target behavior to support situational awareness \cite{dantas2026harbor}, and simulation tools that bring a human operator into the loop of air combat scenarios \cite{silva2025asafg,dantas2025simulation}. In line with human-automation frameworks, such models are best understood as support tools that improve human understanding, consistency, and situational awareness while keeping accountability and control with the human, rather than as autonomous decision-makers \cite{parasuraman2000model,amershi2019guidelines,sharda2018analytics}.
 
Integrating LLMs into military environments, however, raises strict security and reliability challenges. Cloud-based LLM APIs are not compatible with air-gapped networks because of the risk of transmitting sensitive data over public channels \cite{alberts2006understanding}. Furthermore, generic LLMs are prone to ``hallucinations'', meaning they can generate plausible but factually incorrect statements, which is unacceptable where decisions must comply with specific rules of engagement and established doctrine \cite{ji2023survey}. These constraints have motivated the local deployment of smaller, quantized models running entirely on physical servers, combined with grounding mechanisms that restrict responses to verified source materials.
 
Among these grounding mechanisms, RAG reduces hallucination by retrieving document passages semantically similar to the user query and adding them to the model context \cite{lewis2020rag}. While text-only RAG is mature, air operations involve both textual doctrine and visual intelligence products, such as target photographs, structural diagrams, and maps, which motivates multimodal retrieval that aligns visual content with textual context \cite{alayrac2022flamingo,huang2023language}. Importantly, the goal is not to automate image analysis or target recognition, but to assist analysts by retrieving relevant doctrinal criteria and supporting structured reasoning that combines textual procedures with visual observations \cite{deptula2001effects,parasuraman2000model}.
 
Multimodal RAG architectures generally follow two approaches: direct cross-modal embedding models that map images and text into a shared vector space \cite{huang2023language}, or hybrid pipelines in which visual assets are first translated into textual descriptions prior to indexing \cite{alayrac2022flamingo}. For technical manuals, the hybrid approach is frequently preferred, because visual elements such as tables and schematics encode topological relationships that are difficult to capture with generic zero-shot image embeddings. Combining Optical Character Recognition (OCR) with local vision-language captioning converts such diagrams into descriptive text that can be indexed alongside the textual chapters. To improve retrieval precision over single-stage dense search, recent systems add a neural re-ranking stage based on cross-encoders \cite{ji2023survey}; and to align outputs with domain-specific formats without costly fine-tuning, in-context prompt adaptation has emerged as a practical alternative in constrained settings. We build on the hybrid OCR-based approach, neural re-ranking, and prompt adaptation, detailing our implementation in Section~\ref{sec:architecture}.
 
Beyond accuracy, evaluating decision support tools in safety-critical domains also requires methods that capture the human side of the interaction. Recent evidence from multimodal AI applied to aerial monitoring indicates that accuracy alone is insufficient for operational assessment, since models may produce overconfident errors, instruction-following inconsistencies, or unsupported semantic interpretations \cite{dietzsch2026beyond}. This motivates the inclusion of behavioral criteria (such as uncertainty communication, confidence calibration, instruction adherence, and source-grounded reasoning) alongside subjective human-centric metrics. While the former focuses on verifying the AI's output directly (as discussed in Section~\ref{sec:conclusion}), the latter captures the human experience during the interaction. To evaluate this human component, the NASA Task Load Index (NASA-TLX) is a widely used instrument for measuring perceived cognitive workload across operational tasks \cite{hart1988development}, while short instruments such as the Usability Metric for User Experience (UMUX-Lite) provide compact measures of perceived usability \cite{lewis2013umux}. Trust in automation, a key factor in the adoption of AI-based tools, is commonly assessed with short scales, such as the Short Trust in Automation Scale (S-TIAS), derived from established trust frameworks \cite{jian2000foundations,mcgrath2025strust}, and behavioral intention to adopt new technology is often captured through models such as the Unified Theory of Acceptance and Use of Technology (UTAUT) \cite{venkatesh2003user}. Together, these instruments form the basis of the evaluation approach and the full evaluation protocol outlined in Section~\ref{sec:conclusion}.

\section{System Architecture}
\label{sec:architecture}

\begin{figure}[ht]
\centering
\includegraphics[width=\linewidth]{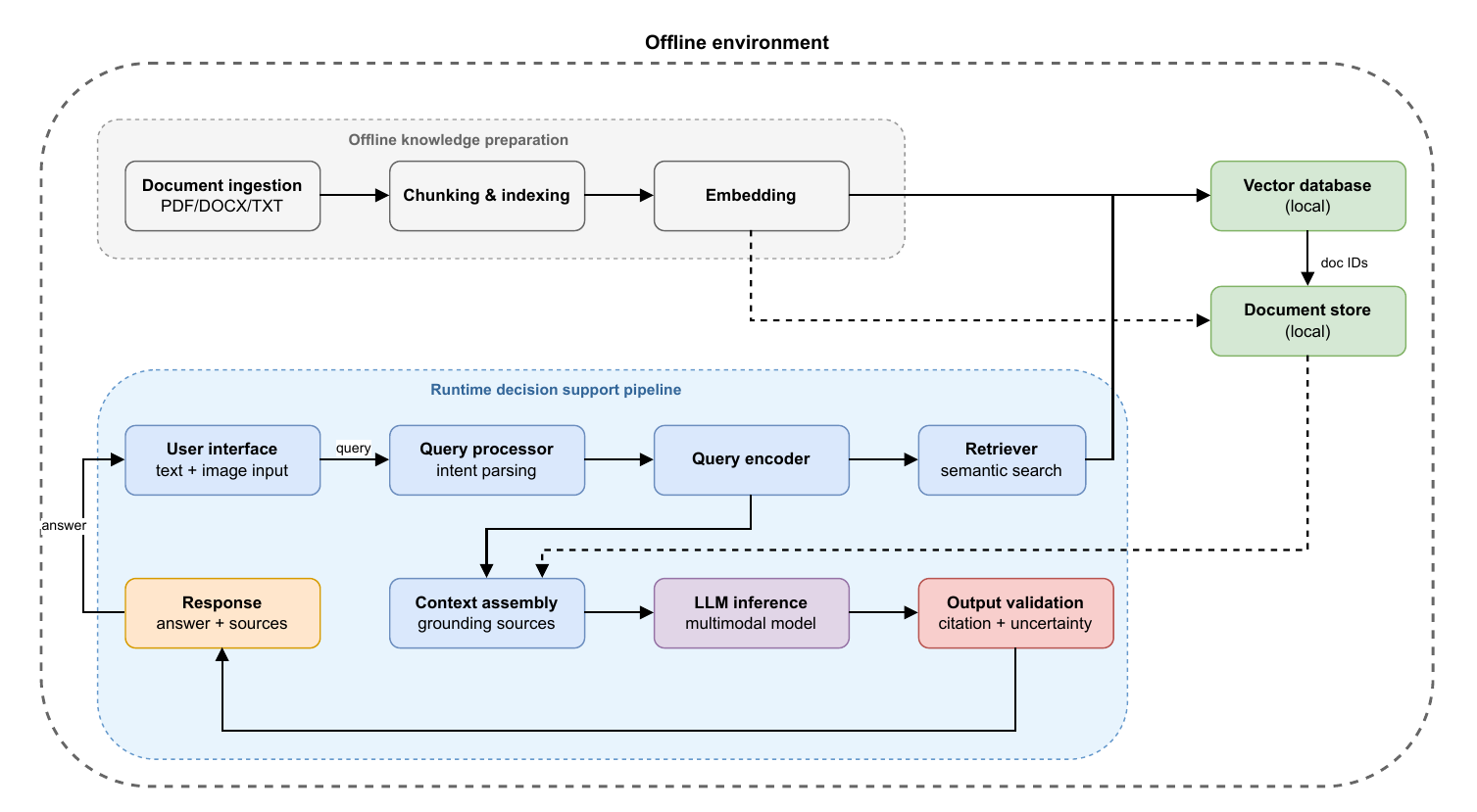}
\caption{Architecture for offline multimodal language model deployment in air operations decision support.}
\label{fig:llm_architecture}
\end{figure}

The proposed decision support system is designed to operate in a strictly disconnected, offline military environment, keeping sensitive data secure while enabling natural language and multimodal interaction with operational documentation. The high-level modular architecture, shown in Figure~\ref{fig:llm_architecture}, consists of two main phases: \textit{offline knowledge preparation} (detailed in Section~\ref{sec:offline_prep}) and \textit{runtime decision support} (comprising retrieval, re-ranking, and user interaction in Sections~\ref{sec:retrieval} and~\ref{sec:runtime_interaction}). This architecture implements a RAG framework, reducing model hallucinations and ensuring that every response generated by the system can be traced back to authoritative doctrinal publications.

\subsection{Offline Document Processing and Indexing}
\label{sec:offline_prep}

The document preparation pipeline processes various technical manuals, regulations, and doctrinal texts (such as PDFs, DOCXs, and TXT files) into a structured knowledge base. To optimize semantic retrieval, documents are processed through a structured pipeline:
\begin{enumerate}
    \item \textbf{Text Segmentation (Chunking):} Documents are parsed and split into overlapping chunks to preserve local semantic context. Based on empirical testing, a chunk size of 1024 tokens with an overlap of 128 tokens is used. This specific configuration balances vector information density with the context window constraints of local LLMs, ensuring that paragraphs and critical tables are not severed across chunks, thereby maintaining semantic coherence.
    \item \textbf{Deduplication:} Chunks are normalized by stripping excess whitespace and filtering duplicates using MD5 hashing. This prevents redundant storage and indexing, reducing the search space and avoiding token waste in the LLM's context window.
    \item \textbf{Multimodal Element Extraction:} Visual elements such as diagrams, tables, and charts are extracted using PyMuPDF \cite{pymupdf}. Rather than discarding these elements, which contain essential schematic and tactical descriptions, the document processing pipeline processes them through a multimodal sub-pipeline. First, OCR is performed via EasyOCR \cite{jaidedai2026easyocr} to extract textual annotations, numbers, and labels embedded within figures. Second, semantic captions are generated locally using a lightweight vision-language model (VLM), MiniCPM-V 8B \cite{yao2024minicpm}, describing system configurations, connections, and topologies. The OCR-extracted text and semantic VLM descriptions are concatenated with their surrounding text context to form unified multimodal chunks, preserving the link between visual assets and text.
    \item \textbf{Vector Indexing:} Text and multimodal chunks are vectorized using a local instance of the multilingual BGE-M3 embedding model (\texttt{BAAI/bge-m3}) \cite{chen-etal-2024-m3}. This model produces 1024-dimensional embeddings, capturing rich semantic relationships across multiple languages, which is essential for translating Portuguese air operations doctrine. The resulting embeddings are indexed in a local, persistent vector database client (ChromaDB) \cite{chromadb} alongside metadata (document title, UUID, page number, and chunk type), ensuring strict traceability.
\end{enumerate}

\subsection{Two-Phase Retrieval and Re-ranking}
\label{sec:retrieval}

To ensure high-precision retrieval under strict latency constraints, a two-phase retrieval pipeline is implemented:
\begin{enumerate}
    \item \textbf{Phase 1: Broad Vector Search (Bi-Encoder Phase):} The user's query is vectorized in real-time using the local BGE-M3 embedding model \cite{chen-etal-2024-m3} and matched against the ChromaDB vector database using cosine similarity. The system retrieves the top-20 candidate chunks ($K_{initial} = 20$). Although bi-encoders are computationally efficient and enable rapid searches over millions of documents, they calculate query and document embeddings independently, which can limit their capacity to capture fine-grained semantic nuances or exact technical manual jargon.
    \item \textbf{Phase 2: Neural Re-ranking (Cross-Encoder Phase):} The query and the retrieved candidate chunks are re-evaluated using a local cross-encoder model, \texttt{BAAI/bge-reranker-v2-m3} \cite{chen-etal-2024-m3}, running with GPU acceleration (or CPU fallback). Unlike the bi-encoder, the cross-encoder processes the query and each candidate chunk concatenated together, performing full joint cross-attention across all tokens. This allows token-to-token comparison, producing a significantly more precise semantic alignment score at the expense of higher latency per pair.
    \item \textbf{Filtering and Constraint:} Chunks with a re-ranking score below a minimum threshold ($S_{min} = 0.1$) are discarded to prevent introducing noise into the prompt. A maximum of the top-5 final chunks ($K_{final} = 5$) is selected to build the context. This constraint prevents overloading the generative LLM's context window, mitigating the risk of the model overlooking relevant information due to ``lost-in-the-middle'' attention degradation.
    \item \textbf{Query Caching:} To minimize computational overhead and response latency, retrieval results are cached in a local in-memory cache with a Time-To-Live (TTL) of 1 hour. This cache is automatically invalidated when new manual versions are uploaded or when documents are modified, guaranteeing that analysts always retrieve the latest authorized tactical guidelines.
\end{enumerate}

\subsection{Runtime Interaction and Feedback Loop}
\label{sec:runtime_interaction}

\begin{figure}[htbp]
\centering
\includegraphics[width=\linewidth]{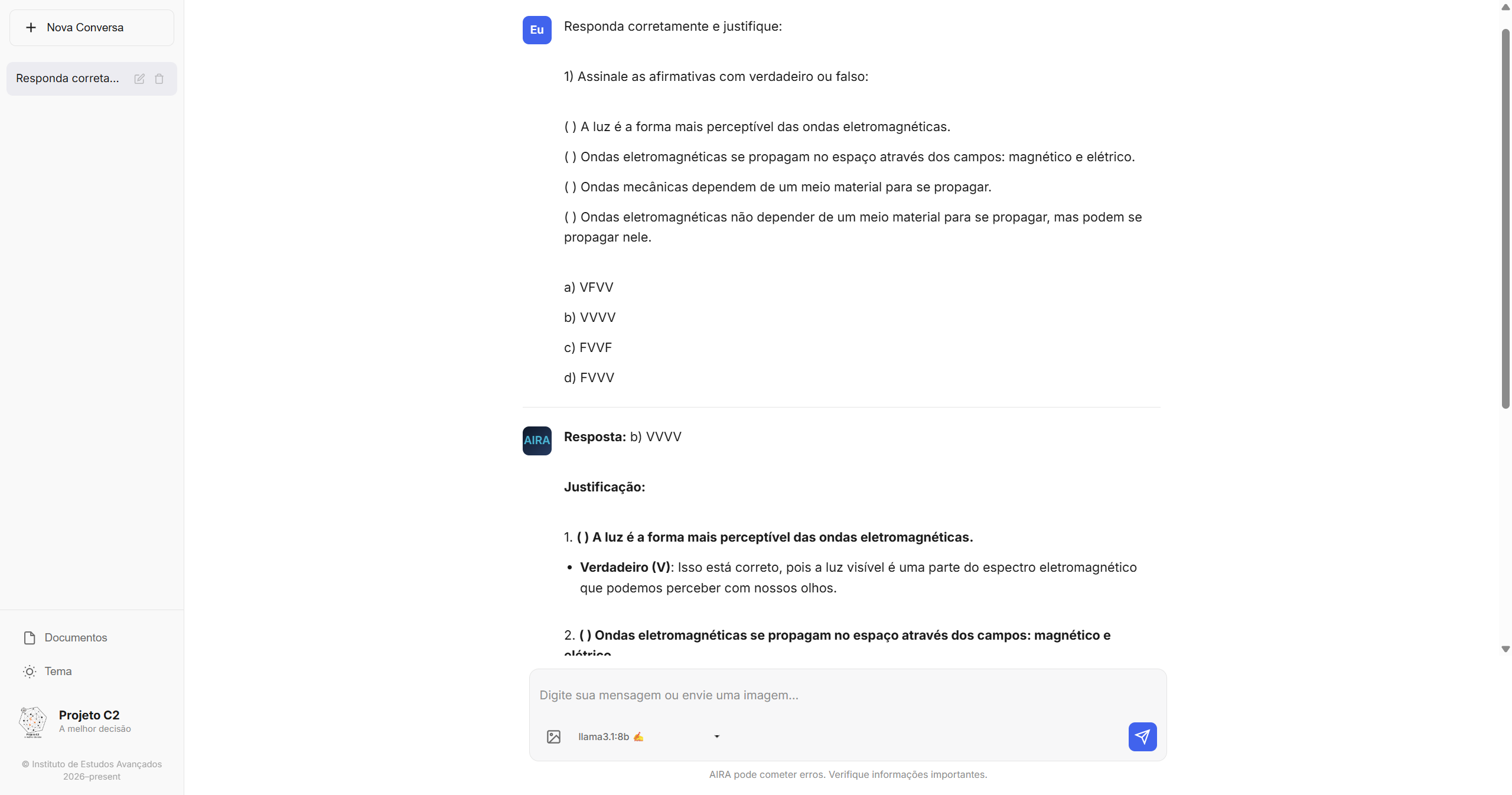}
\caption{User interface mockup illustrating a decision support session where the analyst submits a technical query and receives a structured response grounded in the MCA 200-5 doctrine, complete with source citations.}
\label{fig:mockup}
\end{figure}

During runtime, the system acts as an interactive assistant. The user interacts via a web interface running as a single-page application (SPA) (as illustrated in Figure~\ref{fig:mockup}), submitting queries and optional images to a FastAPI \cite{ramirez2018fastapi} backend server. If the user uploads a reconnaissance image, the backend routes the request to a local multimodal VLM (such as MiniCPM-V 8B \cite{yao2024minicpm} or Moondream run via Ollama \cite{ollama}). For textual queries, the system uses a local generative model, such as Qwen-2.5 14B \cite{qwen2024qwen25}. To provide a highly responsive user experience at the edge, response tokens are streamed back to the frontend in real time using Server-Sent Events (SSE).

The backend dynamically coordinates the retrieval of manual context and builds the system prompt using five distinct layers:
\begin{enumerate}
    \item \textbf{System Identity \& Guidelines:} Global instructions setting the assistant's persona, enforcing strict grounding, and demanding source attribution.
    \item \textbf{Dynamic Learned Rules:} Up to 15 operational constraints compiled dynamically from user feedback to adapt to specific analyst preferences.
    \item \textbf{Retrieved RAG Context:} The top-$K_{final}$ text passages and image caption summaries relevant to the query.
    \item \textbf{Message History:} Up to 100 previous turns to retain conversational context during a multi-turn dialogue.
    \item \textbf{Current Query:} The analyst's latest input, supplemented by raw image data if applicable.
\end{enumerate}

To support continuous workflow alignment, a feedback service captures user thumbs-up/down actions. While positive ratings (thumbs-up) are stored to serve as few-shot exemplars of desired response styles, any comment submitted alongside a negative rating (thumbs-down) is logged and used directly as a prompt correction rule, adapting the model's behavior without resource-intensive fine-tuning. As an illustration of this adaptation mechanism, during system testing a reviewer rated a response containing an unstructured target description with a thumbs-down, submitting the comment: \textit{``Always structure target descriptions using numbered lists to facilitate rapid reading''}. The feedback service automatically registered this comment as an active correction rule, which the prompt builder injected into subsequent queries. Consequently, the generative model began formatting all subsequent target descriptions using structured numbered lists, resolving the layout issue within the next two dialogue turns. This shows that adjusting prompts with feedback rules is a practical way to adapt model outputs in isolated environments without the need for model fine-tuning.

 
\section{Case Study: Electronic Equipment Target Identification}
\label{sec:casestudy}
 
To evaluate the operational utility and cognitive impact of the offline RAG assistant, we designed a case study centered on a highly specialized air operations task: the identification of electronic equipment targets.
 
\subsection{Doctrinal Context: MCA 200-5}
 
In air operations, military analysts must identify and document relevant tactical assets. For electronic systems, this process is governed by the Brazilian Air Force manual \textbf{MCA 200-5 (Manual do Comando da Aeronáutica: Descrição de Alvo, Equipamentos Eletrônicos)} \cite{fabmca2005}. This technical manual establishes visual parameters, frequencies, power metrics, and functional descriptions for electronic targets, including early-warning radars (antennas, command cabins, power generators), height-finder radars (vertical scanning reflectors), and telecommunication and command posts (tower layouts, antenna configurations, feed horns, and signal distribution nodes). Identifying these targets in aerial or satellite reconnaissance imagery requires analysts to map visual features, such as reflector diameters and array configurations, to the classification criteria defined in the manual.
 
\subsection{Participants}
 
The pilot involved $N = 4$ image analysts from the FAB: two officers and two enlisted personnel, all trained in tactical image interpretation (Table~\ref{tab:participant_profile}). Self-reported experience in image/target analysis ranged from 4 to over 7 years, and self-reported familiarity with the MCA 200-5 ranged from moderate to very high. This sample size is representative of the actual operational environment, as the total pool of military analysts who possess the required target-analysis courses, visual credentials, and operational experience to perform electronic-target interpretation is known to be extremely limited. Given this restricted operational setting and the highly specialized nature of the task, the study is framed as a \textbf{pilot evaluation} to characterize baseline behavior and operational feasibility, rather than as a large-scale confirmatory experiment.

\begin{table}[ht!]
\centering
\caption{Profile of the image analysts participating in the pilot study.}
\label{tab:participant_profile}
\resizebox{\columnwidth}{!}{%
\begin{tabular}{|c|c|c|c|}
\hline
\textbf{Participant} & \textbf{Role} & \textbf{Experience in image analysis} & \textbf{Familiarity with MCA 200-5} \\ \hline
Analyst 01 & Officer  & \textgreater 7 years & High      \\ \hline
Analyst 02 & Officer  & 4 -- 7 years          & Moderate  \\ \hline
Analyst 03 & Enlisted & \textgreater 7 years & Very high \\ \hline
Analyst 04 & Enlisted & 4 -- 7 years          & Moderate  \\ \hline
\end{tabular}%
}
\end{table}

\subsection{Pilot Design and Procedure}
\label{sec:pilot_design}

The pilot study was organized into two sequential blocks administered individually to the four FAB image analysts. This initial phase does not yet include the AI-assisted condition for the human participants; instead, it establishes baseline metrics under a manual workflow, which are compared against the offline language models:

\begin{itemize}
    \item \textbf{Block A: Doctrinal Knowledge Assessment.} Analysts individually answered a ten-item multiple-choice test on the MCA 200-5 fundamentals (electromagnetic waves and antennas) consulting only the printed/PDF manual. Start and end times were self-recorded to capture completion time. To explore the operational trade-offs of deploying different local models, the same ten-item assessment was run independently on three LLM configurations (\texttt{qwen2.5:14b}, \texttt{deepseek-r1:14b}, and \texttt{llama3.1:8b}), enabling a direct comparison of accuracy and speed.
    \item \textbf{Block B: Manual REMIR Construction.} Analysts produced a target-reconnaissance report (\textit{Relatório de Missão de Reconhecimento} -- REMIR) for a representative electronic-target image using standard image-analysis tools and the MCA 200-5 manual. Start and end times of the task were self-recorded. Immediately after completing the report, analysts filled out the Raw NASA Task Load Index (RTLX) to characterize the task's cognitive workload.
\end{itemize}

\noindent Since the MCA 200-5 manual and the assessment questions are written in Portuguese, all evaluations (including human tests, model prompts, and RAG outputs) were conducted in Portuguese, testing the models' multilingual capabilities in a specialized domain. The AI-assisted condition, in which analysts interact with the offline RAG assistant, is detailed as part of the future evaluation protocol in Section~\ref{sec:conclusion}.

All model evaluations were conducted locally on a representative tactical edge workstation equipped with an Intel Core Ultra 7 155H CPU, 32~GB of RAM, and an NVIDIA GeForce RTX 4060 Laptop GPU (8~GB VRAM) running Windows 11 and Ollama. Each model was evaluated under two prompt regimes: a \textit{Standard Regime} that required a detailed 2-to-4 line textual justification alongside the answer (essential for auditability), and an \textit{Optimized Regime} that strictly demanded only the letter of the correct alternative (aimed at maximum speed). The standard regime was evaluated using the system's default conversational temperature of 0.3, whereas the optimized regime utilized deterministic greedy decoding (temperature set to 0.0) to maximize consistency on objective multiple-choice selections.
 
\subsection{Evaluation Instruments}
 
This pilot phase relies on two baseline instruments. The first is a \textbf{Doctrinal Knowledge Assessment}: the ten-item test (Block A) measures accuracy (number of correct answers out of ten) and completion time. The second is the \textbf{Raw NASA Task Load Index (RTLX)} \cite{hart1988development}, which captures perceived workload during the manual REMIR-construction task (Block B) across six dimensions: Mental Demand, Physical Demand, Temporal Demand, Performance, Effort, and Frustration, each rated on a 7-point scale. The unweighted (Raw) variant was adopted to reduce administration time; no pairwise weighting procedure was used. In this pilot, the instrument was administered once, referenced to the manual condition, to confirm the operational expectation that manually producing a REMIR is a demanding task and to provide a baseline workload measurement for future comparison with the AI-assisted condition.

\section{Results}
\label{sec:results}

We conducted a pilot exercise with $N = 4$ FAB image analysts (two officers and two enlisted personnel), combining a doctrinal knowledge assessment and a characterization of the cognitive workload associated with manually producing a REMIR under the MCA 200-5. Because of the small, specialized sample, results are reported descriptively (individual scores, means, and standard deviations) and should be interpreted as indicative trends from a pilot study rather than as confirmatory inferential findings.

\subsection{Doctrinal Knowledge Assessment: Human vs. Proposed System Results}
\label{sec:quiz_results}

Table~\ref{tab:quiz} reports the Block~A results for the four FAB analysts on the ten-item MCA 200-5 fundamentals test compared with the proposed system running the \texttt{qwen2.5:14b} model in its optimized fast-response mode, which serves as the baseline configuration (a comparative evaluation of all three candidate models is detailed in Section~\ref{sec:model_eval}). Although the standard regime is essential for auditability and verification, this initial comparison evaluates the system's baseline accuracy and speed under the optimized regime, with trade-offs between execution speed and explainability analyzed in the subsequent section.

\begin{table}[htbp]
\centering
\caption{Knowledge test results: Human analysts vs. Proposed System.}
\label{tab:quiz}
\begin{tabular}{|l|c|c|}
\hline
\textbf{Participant} & \textbf{Score (correct / 10)} & \textbf{Time (min)} \\ \hline
Analyst 01 & 10 & 26.0 \\ \hline
Analyst 02 & 10 & 42.0 \\ \hline
Analyst 03 & 10 & 23.0 \\ \hline
Analyst 04 & 8 & 15.0 \\ \hline
\textbf{Proposed System} & 8 & 7.1 \\ \hline
\end{tabular}
\end{table}

\noindent Three of the four analysts answered all ten items correctly, and the fourth scored 8/10, showing that the underlying doctrinal fundamentals are well established within this group. Completion time, however, varied considerably (15 to 42 minutes, with a mean of 26.5 minutes), reflecting differences in individual search strategy rather than differences in accuracy. The proposed system scored 8/10, matching the score of Analyst 04 and completing the exam in 7.1 minutes, which is faster than all four human analysts. This highlights the capacity of offline RAG architectures to deliver high-quality doctrinal grounding while maintaining competitive execution speeds.

\subsection{Model Evaluation and Prompt Optimization}
\label{sec:model_eval}

\begin{figure}[htbp]
\centering
\includegraphics[width=\linewidth]{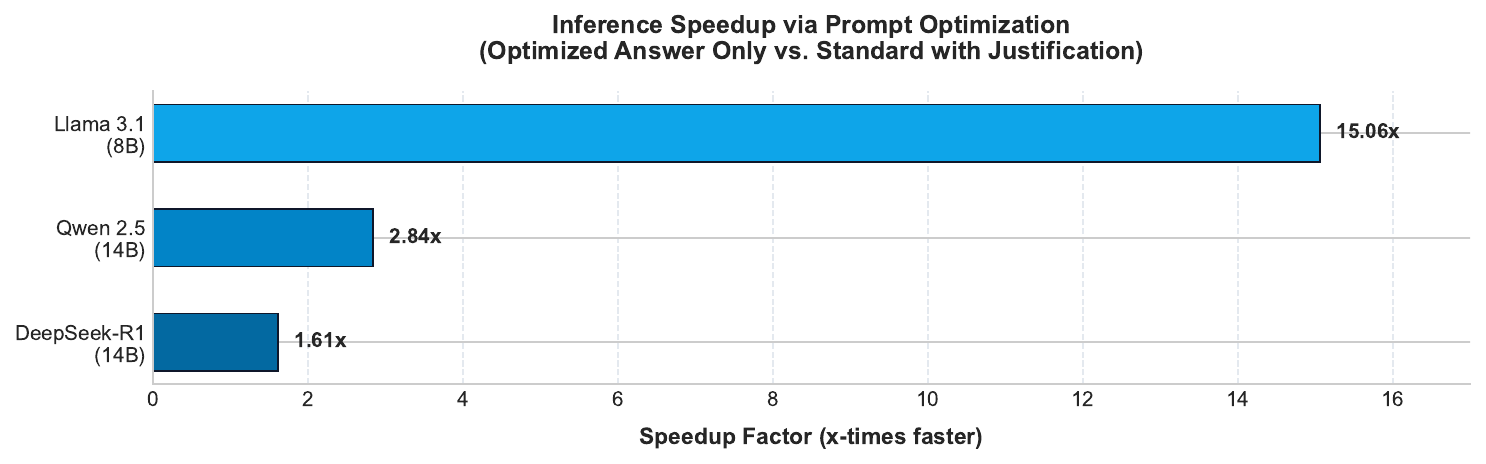}
\caption{Inference acceleration (speedup factor) achieved by the optimized prompting regime relative to the standard prompting regime across the three local models.}
\label{fig:model_comparison}
\end{figure}

The results of the local model evaluation under the different prompting regimes are summarized in Table~\ref{tab:models} and visualized in Figure~\ref{fig:model_comparison}.

\begin{table}[htbp]
\centering
\caption{Local model evaluation under different prompting regimes.}
\label{tab:models}
\begin{tabular}{|l|c|c|c|}
\hline
\textbf{Model \& Regime} & \textbf{Score (correct / 10)} & \textbf{Avg Time (s)} & \textbf{Total Time (min)} \\ \hline
\multicolumn{4}{|l|}{\textit{Standard Regime (with Justification)}} \\ \hline
\texttt{qwen2.5:14b} & 8 & 121.91 & 20.3 \\ \hline
\texttt{deepseek-r1:14b} & 8 & 227.90 & 38.0 \\ \hline
\texttt{llama3.1:8b} & 5 & 44.28 & 7.4 \\ \hline
\multicolumn{4}{|l|}{\textit{Optimized Regime (Answer Only)}} \\ \hline
\texttt{qwen2.5:14b} & 8 & 42.86 & 7.1 \\ \hline
\texttt{deepseek-r1:14b} & 8 & 141.97 & 23.7 \\ \hline
\texttt{llama3.1:8b} & 7 & 2.94 & 0.5 \\ \hline
\end{tabular}
\end{table}

The experimental data reveals that prompt structure is a critical factor for local inference speed. In the standard regime, model response times are highly dominated by the length of the generated response. For the larger 14B models, which exceed the graphics card's memory (8~GB VRAM), processing is heavily constrained by the slow process of transferring the AI model's data back and forth between the computer's main memory (RAM) and the graphics card (VRAM), a process known as memory offloading. Under the optimized regime, the output size is restricted to a single letter option (or limited to internal reasoning steps), bypassing this transfer bottleneck and yielding a substantial speedup across all configurations.

For \texttt{llama3.1:8b}, which fits fully within the graphics card memory, the optimized regime dropped the average retrieval-to-response latency to just 2.94 seconds (a 15x speedup). Interestingly, its accuracy also improved from 5/10 to 7/10. This accuracy gain suggests that smaller models are highly susceptible to reasoning drift or self-distraction when forced to generate verbose justifications before selecting the final option. In an autoregressive sequence (where the AI generates its response word-by-word, building on what it has already written), an early logical error in the justification text tends to bias the model towards selecting an incorrect option to maintain consistency with its own previous text. Bypassing this step keeps the output focused on option selection. While the temperature difference between the regimes (0.3 vs. 0.0, as detailed in Section~\ref{sec:pilot_design}) does not impact execution latency (which depends on prompt-enforced token generation lengths and RAM-to-VRAM transfer overhead), deterministic decoding (greedy decoding, where the model always picks the most likely next word) prevents the sampling of low-probability reasoning paths. Under the optimized regime, the reasoning model \texttt{deepseek-r1:14b} maintained its accuracy of 8/10 but was slower than the other two models under the same regime (averaging 141.97s) due to its mandatory reasoning process (the internal ``chain-of-thought'' generation) which cannot be bypassed at runtime. Overall, \texttt{qwen2.5:14b} represents the most balanced trade-off between accuracy and speed for air operations decision support on constrained edge hardware.

\subsection{Cognitive Workload of the Manual REMIR Task}
\label{sec:tlx_results}

Immediately after manually producing the REMIR for a representative electronic target (Block B), each analyst completed the Raw NASA-TLX. Table~\ref{tab:nasa_tlx} reports the per-dimension means and standard deviations across the four analysts.

\begin{table}[htbp]
\centering
\caption{Subjective workload of the manual REMIR task (Raw NASA-TLX, $N=4$, scale 1--7; SD = Standard Deviation).}
\label{tab:nasa_tlx}
\begin{tabular}{|l|c|}
\hline
\textbf{Workload Dimension} & \textbf{Mean $\pm$ SD} \\ \hline
Mental Demand & 6.0 $\pm$ 1.2 \\ \hline
Physical Demand & 4.0 $\pm$ 1.8 \\ \hline
Temporal Demand & 4.3 $\pm$ 1.5 \\ \hline
Performance & 2.0 $\pm$ 0.8 \\ \hline
Effort & 5.0 $\pm$ 2.2 \\ \hline
Frustration & 4.0 $\pm$ 1.4 \\ \hline
\textbf{Overall (RTLX)} & \textbf{4.2 $\pm$ 1.1} \\ \hline
\end{tabular}
\end{table}

\noindent Mental Demand was the highest-rated dimension (6.0 out of 7, with low variability across analysts), and Effort was also rated highly (5.0 out of 7, but with the largest spread, $SD = 2.2$). The Performance dimension uses a scale where lower values indicate better perceived performance (1 = perfect, 7 = total failure). Consequently, a lower score (such as the observed mean of 2.0) reflects a high sense of success, translating to a low workload contribution. Because a higher numerical score on this scale represents poorer performance (i.e., greater failure and thus higher workload), the dimension naturally aligns with the other five (where higher values represent greater demand and workload). Therefore, the overall RTLX is computed as the direct unweighted mean of all six raw scores. The resulting overall RTLX score (4.2 out of 7) lies above the midpoint of the scale, even for a group with moderate-to-very-high self-reported familiarity with the MCA 200-5 and several years of operational experience. Taken together, these results show that manually cross-referencing visual target features against the MCA 200-5 to produce a REMIR is a cognitively demanding task even for experienced analysts. This strongly supports the case for offline RAG-based decision support; by automating the search and retrieval of complex doctrinal text, the system is positioned to directly reduce mental demand and effort, while maintaining or enhancing the high performance levels (reflected in the low Performance score of 2.0) required in air operations.

\section{Conclusions and Future Work}
\label{sec:conclusion}

This work presented a modular, secure architecture for offline retrieval-augmented language models as decision support tools in air operations. The system combines local text segmentation (1024-token chunks, 128-token overlap), indexing with the multilingual BGE-M3 model, ChromaDB vector storage, neural re-ranking with BGE-Reranker-v2-m3, and dynamic prompt customization through a local user-feedback service, ensuring that every response is traceable to authoritative doctrinal sources while operating entirely offline.

As an initial step toward evaluating this architecture, we conducted a pilot with four FAB image analysts combining a ten-item doctrinal knowledge assessment based on the MCA 200-5 and a Raw NASA-TLX administration referenced to the manual construction of a REMIR. Three of the four analysts answered all ten knowledge items correctly (the fourth scored 8/10), while completion time varied widely (15 to 42 minutes), suggesting that accuracy is largely established within this group but that retrieval efficiency is not. The workload results showed Mental Demand (6.0/7) and Effort (5.0/7) as the highest-rated dimensions, with an overall RTLX of 4.2/7 above the scale midpoint. This shows that, even for experienced analysts, manually cross-referencing visual target features against the MCA 200-5 is a demanding task. Given the small, specialized analyst pool, these results should be read as indicative evidence from a pilot evaluation rather than as confirmatory findings.

These results directly motivate the next phase of this work, for which a within-subjects evaluation protocol has already been designed and is ready for execution:
\begin{enumerate}
    \item \textbf{AI-Assisted Condition:} repeating the REMIR-construction task (Block B) with the same analysts using the offline RAG assistant, with a second Raw NASA-TLX administration to assess the change in cognitive workload relative to the manual baseline reported in Section~\ref{sec:tlx_results}.
    \item \textbf{Perceived Usability, Trust, and Adoption:} administering the UMUX-Lite, S-TIAS, and UTAUT Behavioral Intention instruments (using a 7-point Likert scale), which were introduced in Section~\ref{sec:related}, to assess perceived usefulness and ease of use, trust in the assistant's doctrinal grounding, and intention to adopt the tool operationally.
\end{enumerate}

Beyond this evaluation protocol, future work will focus on five key system-level improvements. First, we plan to develop cross-modal search capabilities by integrating image-matching AI models, such as Contrastive Language-Image Pre-training (CLIP) \cite{radford2021learning} or Sigmoid Language-Image Pre-training (SigLIP) \cite{zhai2023sigmoid}, which connect text and visual data. This will enable direct image-to-image queries, allowing analysts to search for technical diagrams using reconnaissance photos rather than text alone. Second, we will pursue tactical hardware optimization by reducing the model's computational size through quantization and grouping processing tasks through batching, which allows the AI to run efficiently on low-power, portable field servers without requiring high-end workstation hardware. Third, we will introduce automated target description sheets using templates to extract technical details from the analyst's chat logs and compile them directly into standard military reconnaissance reports, eliminating the need to write them manually from scratch. Fourth, we intend to evaluate the porting of this modular architecture to other operational doctrines, such as airspace control or mission planning regulations, to demonstrate its generalizability across different military publications. Fifth, we plan to implement automated verification metrics to mathematically assess faithfulness and relevance, ensuring that the assistant's outputs do not deviate from the retrieved manual text before they are presented to the operator.

\section{Acknowledgments}
This work was supported by the C2 Project --- Command and Control from Massive Aerospace Data Fusion in High-Performance Computing Environments (Agreement No. 01/IEAv/2020), funded by the Department of Aerospace Science and Technology (DCTA), Brazilian Air Force. The authors would also like to thank the anonymous military analysts who volunteered to participate in this study.

\section{Contact Author Email Address}
\small Joao P. A. Dantas: \underline{\href{mailto:jpdantas@ita.br}{jpdantas@ita.br}}

\section{Copyright Statement}
\begin{small}
The authors confirm that they, and/or their organization, hold copyright on all original material included in this paper and have obtained permission for any third-party material. Permission is granted for publication and distribution as part of the ICAS proceedings.
\end{small}

\biblio{bibliography}
\label{bibliography}

\end{document}